\documentclass[10pt,twocolumn,letterpaper]{article}

\usepackage[pagenumbers]{cvpr} 

\usepackage{graphicx}
\usepackage{amsmath}
\usepackage{amssymb}
\usepackage{booktabs}

\usepackage{rotating}
\usepackage{url}
\usepackage{mathtools}
\usepackage{caption}
\usepackage{subcaption}
\usepackage{multirow}
\usepackage{multicol}
\usepackage{makecell}
\newcommand{\xmark}{\ding{55}}%
\usepackage{pifont}
\usepackage{xcolor}
\usepackage{pdfrender}

\usepackage{xcolor,colortbl}
\definecolor{Gray}{gray}{0.95}

\newcommand{\stdv}[1]{\scriptsize$\pm$#1}
\newcommand*{\boldcheckmark}{%
  \textpdfrender{
    TextRenderingMode=FillStroke,
    LineWidth=.5pt, 
  }{\checkmark}%
}
\newcommand*{\boldxmark}{%
  \textpdfrender{
    TextRenderingMode=FillStroke,
    LineWidth=.5pt, 
  }{\xmark}%
}
\newcommand*{\affmark}[1][*]{\textsuperscript{#1}}
\usepackage[pagebackref,breaklinks,colorlinks]{hyperref}

\usepackage[capitalize]{cleveref}
\crefname{section}{Sec.}{Secs.}
\Crefname{section}{Section}{Sections}
\Crefname{table}{Table}{Tables}
\crefname{table}{Tab.}{Tabs.}

\def\cvprPaperID{5006} 
\def\confName{CVPR}
\def\confYear{2023}

\begin{document}

\title{Improving Evaluation of Debiasing in Image Classification}

\author{
Jungsoo Lee\affmark[1] \; Juyoung Lee\affmark[2] \; Sanghun Jung\affmark[1] \; Jaegul Choo\affmark[1]\vspace{0.2cm}\\
\affmark[1]KAIST AI \; \affmark[2]Kakao Enterprise\vspace{0.2cm}\\
\texttt{\footnotesize\affmark[1]\{bebeto, shjung13, jchoo\}@kaist.ac.kr} \; \texttt{\footnotesize\affmark[2]michael.jy@kakaoenterprise.com}\\
}
\maketitle

\begin{abstract}
Image classifiers often rely overly on peripheral attributes that have a strong correlation with the target class (\textit{i.e.,} dataset bias) when making predictions.
Due to the dataset bias, the model correctly classifies data samples including bias attributes (\textit{i.e.,} bias-aligned samples) while failing to correctly predict those without bias attributes (\textit{i.e.,} bias-conflicting samples).
Recently, a myriad of studies focus on mitigating such dataset bias, the task of which is referred to as \textbf{debiasing}. 
However, our comprehensive study indicates several issues need to be improved when conducting evaluation of debiasing in image classification.
First, most of the previous studies do not specify how they select their hyper-parameters and model checkpoints (\textit{i.e.,} tuning criterion). 
Second, the debiasing studies until now evaluated their proposed methods on datasets with excessively high bias-severities, showing degraded performance on datasets with low bias severity.  
Third, the debiasing studies do not share consistent experimental settings (\textit{e.g.,} datasets and neural networks) which need to be standardized for fair comparisons. 
Based on such issues, this paper 1) proposes an evaluation metric `Align-Conflict (AC) score' for the tuning criterion, 2) includes experimental settings with low bias severity and shows that they are yet to be explored, and 3) unifies the standardized experimental settings to promote fair comparisons between debiasing methods.  
We believe that our findings and lessons inspire future researchers in debiasing to further push state-of-the-art performances with fair comparisons.
\end{abstract}

\section{Introduction}
\label{sec: intro}
\begin{figure}[t!]
    \centering
    \includegraphics[width=0.5\textwidth, clip]{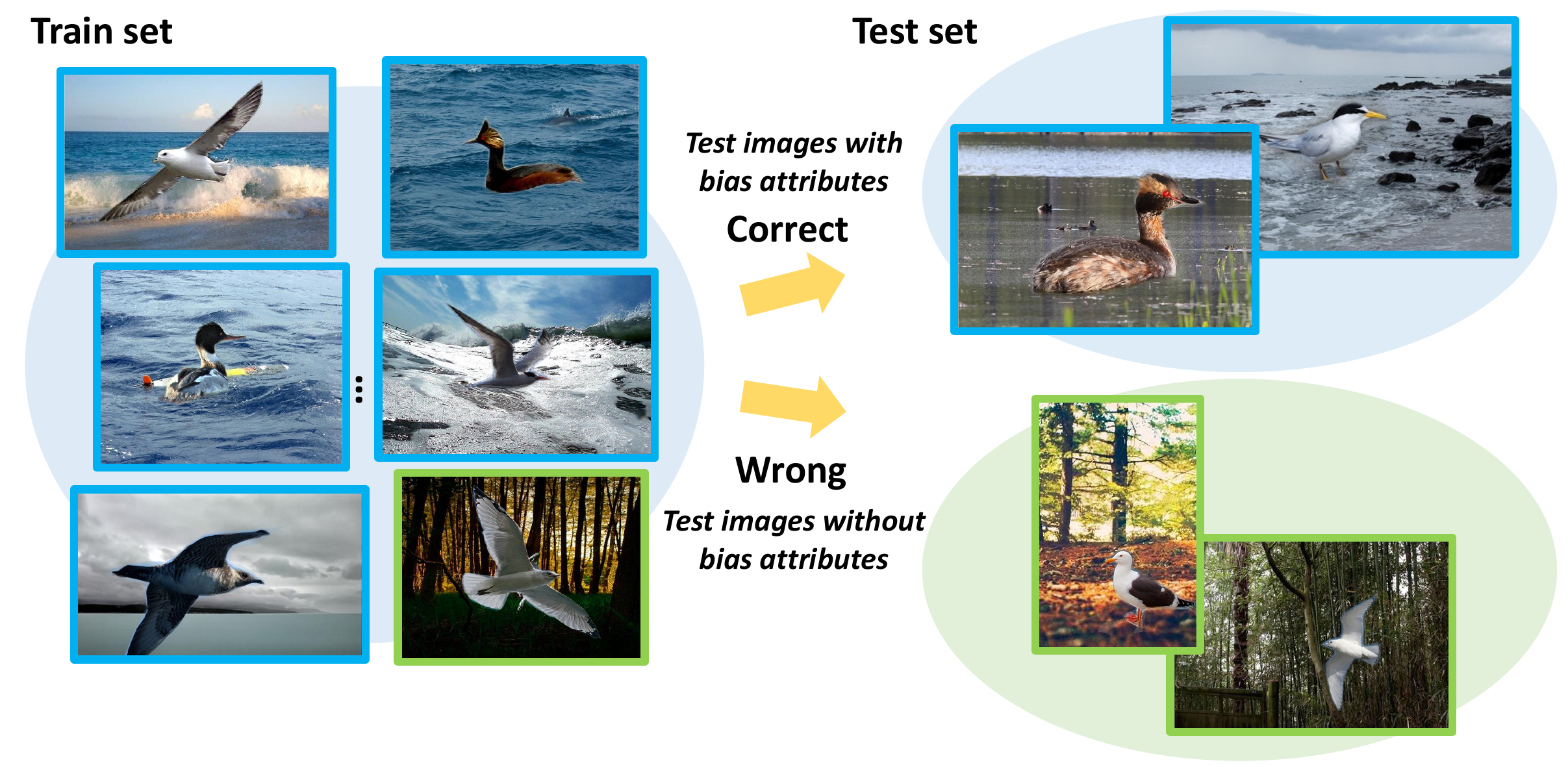}
    \caption{Description of the debiasing task. When trained with a biased dataset, the image classifier relies on the bias attribute (\textit{e.g.,} water background) as the visual cue to predict the target class (\textit{e.g.,} waterbirds). This leads the model to misclassify test images without the bias attribute such as waterbirds in the land background. The blue and green bordered images indicate the images with bias attributes and the ones without bias attributes, respectively.}
    \label{fig: teaser}
    \vspace{-0.5cm}
\end{figure}

Image classification models tend to heavily rely on the correlation between peripheral attributes and the target class, which is termed as \textit{dataset bias}~\cite{unbiaslook2011torralba,nam2020learning,disentangled}.
Such dataset bias is found in datasets where the majority of images include visual attributes that frequently co-occur with the target class but do not innately define it, referred to as \textit{bias attributes}. 
For example, as shown in Fig.~\ref{fig: teaser}, a training set of waterbird images may include the majority of waterbirds in the water background (\textit{i.e.,} bias-aligned samples) while waterbirds can also be found on lands (\textit{i.e.,} bias-conflicting samples). 
In such a case, the image classifier may use the water background as the visual cue for predicting the waterbird images while not learning the \textit{intrinsic attributes}, the visual attributes which innately define a certain class (\textit{e.g.,} beaks and wings of waterbirds).  
This becomes problematic during the inference stage since the classification model may misclassify waterbird images without the water background.
Instead, it may also wrongly predict images with other classes taken in the water background (\textit{e.g.,} landbirds in the water) as waterbirds since it heavily relies on the strong correlation between the water background (bias attribute) and the bird type (target class). 
The task of \textit{debiasing} aims to mitigate such dataset bias.

\begin{table*}[t!]
\centering
\scalebox{0.825}{
\begin{tabular}{c|ccccccccc}
\toprule
Algorithm & Tuning Criterion & \makecell{Lowest \\ Bias severity} & Dataset & Intrinsic attr. & Bias attr. & Architecture & 
\\
\midrule
\multirow{2}{*}{\makecell{ReBias~\cite{bahng2019rebias}}} & \multirow{2}{*}{\makecell{Not specified}} &  1pct & Colored MNIST & Digits & Color & Conv layers 
\\
& & N.A & ImageNet & ImageNet classes & Texture & ResNet18 (init.) 
\\
\midrule
\multirow{2}{*}{\makecell{EnD~\cite{EnD}}} 
& \multirow{2}{*}{\makecell{Preprocessed Valid.}}  & 1pct & Colored MNIST & Digits & Color & Conv layers 
\\
 & & 5pct & CelebA & Facial attributes & Gender & ResNet18 (pretr.) 
\\
\midrule
\multirow{4}{*}{\makecell{LfF~\cite{nam2020learning}}}  & \multirow{4}{*}{\makecell{Not specified}} & 5pct & Colored MNIST & Digits & Color & MLP \\
& & 5pct & Corrupted CIFAR10 & CIFAR10 classes & Corruption & ResNet20 (init.) & & 
\\
& & 5pct & CelebA & Facial attributes & Gender & ResNet18 (pretr.) & & 
\\
& & - & BAR & Action & Background & ResNet18 (pretr.) & & 
\\
\midrule
\multirow{3}{*}{\makecell{DisEnt~\cite{disentangled}}} 
& \multirow{3}{*}{\makecell{Not specified}} & 5pct & Colored MNIST & Digits & Color & MLP  
\\
& & 5pct & Corrupted CIFAR10 & CIFAR10 classes & Corruption & ResNet18 (init.) & &
\\
& & 0.5pct & BFFHQ & Age & Gender & ResNet18 (init.) & &
\\
\midrule
\bottomrule
\end{tabular}
}
\caption{Summary of the experimental settings of existing debiasing studies. We include 1) specification of the tuning criterion, 2) the lowest bias severity of each dataset, and 3) different datasets along with bias and intrinsic attributes and neural network architectures. Pretr. and init. indicates that the model is pretrained with ImageNet~\cite{imagenet} and it is randomly initialized, respectively. Valid. refers to the validation set. Note that previous studies either do not specify how their tuning criterion or preprocess the validation set to follow the data distribution of the test set which requires annotations of bias-conflicting samples for the training set.}
\vspace{-0.3cm}
\label{tab:inconsistent}
\end{table*}

Until recently, numerous debiasing methods have been proposed~\cite{li2019repair,towards_fairness,women_snowboard,bahng2019rebias,LNL,nam2020learning,disentangled,BiasEnsemble}.
However, we found several issues that need to be improved when evaluating debiasing methods in image classification.
Table~\ref{tab:inconsistent} summarizes the experimental settings of the recent debiasing studies. 
First, the previous studies either do not specify their tuning criterion~\cite{bahng2019rebias, nam2020learning, disentangled} or construct the validation set to have an equal number of bias-aligned samples and bias-conflicting samples~\cite{EnD}. 
Throughout the paper, tuning criterion refers to selecting 1) hyper-parameters and 2) model checkpoints used during the test phase.
Tuning criterion in debiasing is a challenging task due to the different data distributions between the validation set and the test set. 
To be more specific, the validation set is a \emph{biased} set which mainly consists of bias-aligned samples when they are randomly sampled from the training set. 
On the other hand, the test set used for evaluating debiasing methods is an \emph{unbiased} set. 
Due to such a discrepancy, simply selecting model checkpoints using the average accuracy of the validation set reflects the performance of bias-aligned samples mainly, showing poor performance on the unbiased test set (Table~\ref{tab:main_difference}). 
One possible approach is to preprocess the validation set to have a similar data distribution to the test set~\cite{EnD}.
However, such an approach has the following two issues.
The first issue is that we need to know whether a given \emph{training} sample is a bias-aligned sample or a bias-conflicting sample in order to adjust the number in the validation set.
However, recent debiasing methods~\cite{nam2020learning, disentangled, BiasEnsemble} assume that such information is not available during training due to the high labor intensity required for such annotations. 
Another issue is that an additional number of bias-conflicting samples are required to be sampled from the training set in order to adjust the number which reduces the sources of information the model can learn from. 
Therefore, rather than preprocessing the validation set, we need an evaluation metric which reflects the data distribution of the test set by using the validation set composed of randomly sampled training samples.

Second, the previous debiasing studies only conducted evaluations with high bias severities.
In other words, the training datasets used in previous studies included an excessively small number of bias-conflicting samples. 
`Bias severity' in Table~\ref{tab:inconsistent} indicates the lowest bias severity conducted for each dataset. 
We observe that the previous studies focused on improving debiasing performances assuming that the training dataset consists of only up to 5 percent of bias-conflicting samples.
In the real world, however, not all biased datasets have high bias severity.
A non-trivial number of bias conflicting samples may be included in the biased dataset (\textit{i.e.,} low bias severity). 
In such an aspect, evaluation with low biased severity is also required. 
In fact, we observe that existing debiasing methods fail to consistently outperform the vanilla method (\textit{i.e.,} training without debiasing module) with datasets of low bias severity (Table~\ref{tab:main_f1_acc} and Table~\ref{tab:main_f1_conflict}), leaving room for the future debiasing researchers to improve with.  
Third, we found that the previous studies do not share consistent experimental settings including datasets and neural network architectures. 
We believe that a standardized experimental setting is required to further encourage fair comparisons between debiasing studies for future researchers.

In this work, we propose experimental settings which improve the evaluation of debiasing in image classification. 
First, we propose `Align-Conflict (AC) score' for the tuning criterion in order to take account of both performances on bias-aligned samples and bias-conflicting ones. 
We compute the harmonic mean between the accuracy of bias-aligned samples and that of bias-conflicting samples in the validation set and select hyper-parameters and model checkpoints that achieve the best AC score of the validation set.
Second, we reveal that the recent state-of-the-art debiasing methods~\cite{nam2020learning, disentangled, BiasEnsemble}, those which do not 1) utilize explicit bias labels and 2) predefine bias types (\textit{e.g.,} color), fail to consistently outperform the vanilla method in datasets with low bias severity. 
Third, we unify the neural network architectures and datasets to construct a consistent experimental setting for making fair comparisons between debiasing studies. 

Our main contributions are summarized as follows: 
    \vspace{-0.2cm}
\begin{itemize}
    \item We propose a tuning criterion for debiasing studies termed as `Align-Conflict (AC) score' which takes account of both bias-aligned samples and bias-conflicting ones.  
    \vspace{-0.2cm}
    \item We reveal that the recent debiasing studies fail to consistently outperform the vanilla method on datasets with low bias severity which future debiasing studies may improve with. 
    \vspace{-0.2cm}
    \item We unify the inconsistent experimental settings to promote fair comparisons between debiasing methods.
\end{itemize}

\section{Related Work}
\subsection{Debiasing in Image Classification}
Early studies in debiasing define bias types in advance~\cite{li2019repair, bahng2019rebias, wang2018hex, agrawal2016analyzing} or even use the bias labels for training~\cite{sagawa2019distributionally, EnD, LNL}. 
Acquiring bias labels, however, may be challenging and labor intensive since it necessitates manual labeling by human annotators with sufficient understanding of the underlying bias of a given dataset~\cite{nam2020learning}. 
While not using the bias labels during training, Bahng~\textit{et al.} propose ReBias which defines a bias type in advance and leverages neural network architectures tailored for the predefined bias type (\textit{e.g.,} using models of small receptive fields for addressing color and texture bias)~\cite{bahng2019rebias}.
Still, predefining a bias type 1) limits the debiasing capacity in other bias types especially when they are unknown and 2) requires human labor to manually identify the bias types~\cite{disentangled}. 

To address such an issue, recent debiasing methods do not utilize bias labels or predefine bias types when mitigating the dataset bias~\cite{nam2020learning, disentangled, BiasEnsemble}. 
Nam~\textit{et al.} propose LfF which emphasizes the bias-conflicting samples to mitigate the dataset bias by identifying them based on the intuition that bias is easy to learn~\cite{nam2020learning}. 
Under the same assumption, Lee~\textit{et al.} propose a feature-level augmentation method to augment the bias-conflicting samples at the feature level~\cite{disentangled}. 
Training a debiased classifier without utilizing explicit bias labels or predefined bias types is challenging but practical, so recent studies aim to follow such a training setting.

\subsection{Standardizing inconsistent experimental settings}
A plethora of algorithms in domain generalization \cite{learning-single, cha2021swad, zhou2021mixstyle, domainbed, kim2021selfreg, dg-unified}, metric learning \cite{soft-triplet, metric-learning-hardaware, metric_learning_ranking, metric_learning_adversarial}, and cross-domain few-shot learning \cite{crossdomain-fewshot-museum, cross_domain_feature, cross_domain_oneshot} have inconsistent experimental settings which use different datasets and neural network architectures.
Due to this fact, researchers in each field build benchmark repositories and conduct fair comparisons via extensive experiments. 
For example, Gulrajani~\textit{et al.} build DomainBed which points out that domain generalization methods which do not specify model selection criterion may have used the test set to select parameters.~\cite{domainbed}. 
Also, Musgrave~\textit{et al.} similarly propose an evaluation protocol in metric learning to fix the flaws and problems of unfair comparisons found in the existing metric learning studies~\cite{metric-reality}.
Guo~\textit{et al.} reveal that the recent state-of-the-art cross-domain few-shot learning methods show poor performance when evaluated with datasets containing large domain discrepancy by building a benchmark named Broader Study of Cross-Domain Few-Shot Learning~\cite{cross-domain}.
Such studies received appreciation for their contribution to standardizing experimental settings and promoting fair comparisons between proposed methods. 

In the meanwhile, there exist studies which promote fair comparisons between debiasing methods~\cite{benchmark_cvpr, benchmark_neurips}.
Wang~\textit{et al.} propose a benchmark for debiasing in visual recognition including four debiasing methods and two different datasets, and provide an analysis of their experiments (\textit{e.g.,} revealing the shortcoming of adversarial approach on bias mitigation)~\cite{benchmark_cvpr}.
Reddy~\textit{et al.} propose a benchmark which compares the performance of vanilla method (\textit{e.g.,} MLP and CNN) optimized with diverse fairness-related metrics~\cite{benchmark_neurips}.
While our work is similar to those benchmark studies in that we compare existing debiasing studies under fair evaluation settings, we solve the following issues that they do not consider.
First, we propose a tuning criterion specialized for debiasing which was not discussed in the previous benchmarks. 
Second, we include the low bias severity in the experimental setting based on the finding that recent debiasing methods fail to consistently outperform the vanilla method with low bias severity. 
We believe following our experimental settings improves the evaluation of debiasing in image classification. 

\section{Preliminaries}
Inspired by the work of Bahng~\textit{et al.}~\cite{bahng2019rebias}, we formulate the debiasing task.
In a biased dataset, an input $X_{train}$ with the target class $Y_{train}$ includes the intrinsic attributes $I_{train}$ and the bias attributes $B_{train}$.
In the case of aforementioned waterbird images in the water background example, the wings and beaks of waterbirds correspond to $I_{train}$ while the water background corresponds to $B_{train}$. 
The difference between $I$ and $B$ is that $I$ always appears in the target class while $B$ does not necessarily appear since it does not innately define the target class.

In supervised learning, a classifier $f$ can output unintended results under different circumstances~\cite{shortcut}. 
To be more specific, $f$ which is trained with the biased dataset learns the unwanted correlation between $B_{train}$ and $Y_{train}$ rather than learning $I_{train}$.
$f$ may even rely on $B_{train}$ to predict $Y_{train}$.
Learning of such spurious correlations shows significant performance drop when $B_{test} \neq B_{train}$, meaning that $X_{test}$ does not necessarily include $B_{train}$.
$f$ is prone to misclassifying $X_{test}$ especially when $B_{train}$ is included in other target classes (\textit{e.g.,} wrongly predicting landbirds in the water background as waterbirds).
Therefore, the debiasing task requires $f$ to learn $I$ of a target class regardless of $B_{train}$ since $I_{train}$=$I_{test}$ for a given target class.
Following the work of Nam~\textit{et al.}~\cite{nam2020learning}, we define \textit{bias-aligned} samples as $X$ including $B_{train}$ (\textit{e.g.,} waterbirds in the water background) and \textit{bias-conflicting} samples as $X$ without $B_{train}$ (\textit{e.g.,} waterbirds on land).


\section{Align-Conflict score}
\label{sec:ac_f1_score}

Assuming that the validation set mainly consists of bias-aligned samples similar to the training set, selecting model checkpoints based on the average accuracy of the validation set fails to achieve a reasonable performance on bias-conflicting samples (Table~\ref{tab:main_difference}). 
This is due to the fact that the averaged accuracy mainly reflects the predictions on bias-aligned samples. 
In order to solve this issue, one may preprocess the validation set to have an equal number of bias-aligned ones and bias-conflicting ones by sampling them from the training set~\cite{EnD}.
However, as aforementioned, such an approach has two issues.
First, in order to select an equal number of bias-aligned samples and bias-conflicting samples from the training set, we need to know whether a given training sample is a bias-aligned one or a bias-conflicting one. 
However, acquiring such information on the training set is labor intensive, so recent debiasing methods assume that we do not have such labels on the training samples~\cite{nam2020learning, disentangled, BiasEnsemble}. 
Second, in order to adjust the number of bias-conflicting samples to the number of bias-aligned ones, we need to sample an additional number of bias-conflicting ones from the training set but such sampling reduces the sources of information the model can learn from.



To this end, we propose a novel metric termed `Align-Conflict (AC) score' which takes account of performance on both bias-aligned samples and bias-conflicting ones for the tuning criterion. 
Given a biased validation set, we compute the harmonic mean of the accuracy of bias-aligned samples and that of bias-conflicting ones. 
AC score is formulated as 
\begin{equation}
    \text{AC score} = \frac{2\cdot\frac{n_A}{N_A}\cdot\frac{n_C}{N_C}}{\frac{n_A}{N_A}+\frac{n_C}{N_C}},
\end{equation}
where $n_A$ and $n_C$ indicate the number of correctly classified bias-aligned samples and bias-conflicting samples, respectively, while $N_A$ and $N_C$ indicate the total number of bias-aligned samples and bias-conflicting samples, respectively, in the validation set. 
To the best of our knowledge, our work is the first work in debiasing studies to propose an evaluation metric which selects the tuning criterion based on the performance of both bias-aligned samples and bias-conflicting samples without preprocessing the validation set. 

\vspace{-0.2cm}
\section{Experiments}
\subsection{Debiasing Methods}
\label{sec:debiasing_method}
We conducted extensive experiments with 9 debiasing methods: Vanilla, Learning Not to Learn (\textbf{LNL}~\cite{LNL}), Entangling and Disentangling deep representations for bias correction (\textbf{EnD}~\cite{EnD}), Projecting Superficial Statistics (\textbf{Hex}~\cite{wang2018hex}),  Learning Debiased Representations with Biased Representations (\textbf{ReBias}~\cite{bahng2019rebias}), Soft Bias-Contrastive Loss (\textbf{Softcon}~\cite{softcon}), Learning from Failure (\textbf{LfF}~\cite{nam2020learning}), Learning Debiased Representation via Disentangled Feature Augmentation (\textbf{DisEnt}~\cite{disentangled}), BiasEnsemble (\textbf{BiasEnsemble}~\cite{BiasEnsemble}). 
Further details of each debiasing algorithm are included in the Supplementary. 

\subsection{Dataset}
\label{sec:dataset}
We use four datasets: Colored MNIST~\cite{disentangled}, biased FFHQ (BFFHQ)~\cite{biaswap}, Waterbirds~\cite{sagawa2019distributionally}, and biased action recognition (BAR)~\cite{nam2020learning}. 
The training set and the validation set for each dataset have various ratios of bias-conflicting samples in order to evaluate algorithms under varying levels of bias severity.
The ratios are 0.5\%, 1\%, 2\%, 5\%, and 20\% for Colored MNIST, BFFHQ, and Waterbirds, and 1\%, 5\%, and 20\% for BAR.
We use the unbiased test set for evaluating methods.
As aforementioned in Table~\ref{tab:inconsistent}, previous studies only conducted experiments under high bias severity, but we also include datasets with low bias severity (ratio of 20\%).
Further details of datasets are included in our Supplementary.

\begin{table*}[t!]
\centering
\scalebox{0.85}{
\begin{tabular}{ccccccccccccc}
\toprule
\multirow{2.5}{*}{Dataset} 
& \multirow{2.5}{*}{Ratio (\%)} 
&
& Vanilla
& LNL~\cite{LNL}
& EnD~\cite{EnD}
& HEX~\cite{wang2018hex}
& ReBias~\cite{bahng2019rebias}
& Softcon~\cite{softcon}
& LfF~\cite{nam2020learning}
& DisEnt~\cite{disentangled}
& BiasEnsemble~\cite{BiasEnsemble}

\\
\cmidrule{4-12}
&&
& \textcolor{black}{\boldxmark} \textcolor{black}{\boldxmark}               
& \textcolor{black}{\boldcheckmark} \textcolor{black}{\boldcheckmark}       
& \textcolor{black}{\boldcheckmark} \textcolor{black}{\boldcheckmark}       
& \textcolor{black}{\boldxmark} \textcolor{black}{\boldcheckmark}           
& \textcolor{black}{\boldxmark} \textcolor{black}{\boldcheckmark}           
& \textcolor{black}{\boldxmark} \textcolor{black}{\boldcheckmark}           
& \textcolor{black}{\boldxmark} \textcolor{black}{\boldxmark}               
& \textcolor{black}{\boldxmark} \textcolor{black}{\boldxmark}               
& \textcolor{black}{\boldxmark} \textcolor{black}{\boldxmark}               
\\
\midrule
\multirow{5}{*}{\makecell{Colored \\ MNIST}} 
& 0.5 & & 34.61 & 34.54 & 36.69 & 24.58 & 21.62 & 31.29 & 62.01 & 59.99 & \underline{\textbf{66.71}} \\ & 1.0 & & 49.87 & 46.83 & 57.87 & 39.80 & 31.01 & 47.66 & 76.17 & 73.30 & \underline{\textbf{77.94}} \\ & 2.0 & & 66.84 & 61.35 & 70.71 & 50.76 & 47.04 & 60.79 & 82.63 & 76.86 & \underline{\textbf{82.98}} \\ & 5.0 & & 81.56 & 78.56 & 81.50 & 79.58 & 73.24 & 79.46 & 86.43 & 84.05 & \underline{\textbf{88.74}} \\ & 20.0 & & \underline{\textbf{92.77}} & 90.73 & 92.22 & 91.02 & 91.68 & 89.57 & 90.76 & 91.97 & 91.32 \\ 
\midrule
\multirow{5}{*}{\makecell{BFFHQ}} 
& 0.5 & & \underline{73.46} & 76.20 & 75.41 & 76.09 & \textbf{76.64} & 76.09 & 73.07 & 71.72 & 69.61 \\ & 1.0 & & 78.05 & 78.16 & 79.51 & 79.69 & \textbf{80.98} & 78.75 & 73.37 & 76.10 & \underline{78.98} \\ & 2.0 & & \underline{81.47} & 83.93 & 82.42 & 83.40 & \textbf{86.31} & 83.55 & 77.22 & 79.24 & 80.09 \\ & 5.0 & & \underline{89.42} & 90.38 & 88.37 & 90.00 & \textbf{91.91} & 90.65 & 78.55 & 85.10 & 79.80 \\ & 20.0 & & \underline{95.47} & 95.75 & 95.66 & 95.22 & \textbf{96.43} & 95.47 & 83.59 & 91.04 & 89.60 \\
\midrule
\multirow{5}{*}{\makecell{Waterbirds}} 
& 0.5 & & 64.18 & 65.22 & 65.52 & 70.09 & 66.57 & \textbf{77.96} & 59.80 & \underline{66.55} & 61.73 \\ & 1.0 & & 64.28 & 66.69 & 68.37 & 76.31 & 65.16 & \textbf{77.72} & 61.69 & \underline{66.21} & 61.34 \\ & 2.0 & & \underline{68.87} & 67.29 & 77.02 & 74.08 & 67.88 & \textbf{78.03} & 66.08 & 53.86 & 60.64 \\ & 5.0 & & 73.24 & 70.66 & 74.87 & 74.40 & 70.95 & \textbf{77.97} & 65.84 & \underline{74.35} & 70.84 \\ & 20.0 & & \underline{79.23} & \textbf{79.74} & 79.54 & 79.40 & 79.30 & 77.90 & 72.94 & 77.22 & 77.95 \\ 
\midrule
\multirow{3}{*}{\makecell{BAR}} 
& 1.0 & & 81.36 & - & - & 80.64 & 81.89 & 77.14 & 82.11 & \underline{\textbf{82.82}} & 82.46 \\ & 5.0 & & 90.86 & - & - & 88.39 & 89.07 & 86.50 & 90.71 & 91.25 & \underline{\textbf{91.46}} \\ & 20.0 & & 95.05 & - & - & 93.38 & 93.59 & 90.35 & \underline{\textbf{95.61}} & 95.61 & 95.05 \\ 
\midrule
\bottomrule
\end{tabular}
}
\vspace{-0.2cm}
\caption{Image classification accuracy evaluated on unbiased tests sets. We selected hyper-parameters and model checkpoints using AC score. The \textit{cross} and \textit{check} marks denote whether each algorithm requires 1) bias labels during training and 2) prior knowledge on bias types.}
\label{tab:main_f1_acc}
\end{table*}

\begin{table*}[t!]
\centering
\scalebox{0.85}{
\begin{tabular}{ccccccccccccc}
\toprule
\multirow{2.5}{*}{Dataset} 
& \multirow{2.5}{*}{Ratio (\%)} 
&
& Vanilla
& LNL~\cite{LNL}
& EnD~\cite{EnD}
& HEX~\cite{wang2018hex}
& ReBias~\cite{bahng2019rebias}
& Softcon~\cite{softcon}
& LfF~\cite{nam2020learning}
& DisEnt~\cite{disentangled}
& BiasEnsemble~\cite{BiasEnsemble}

\\
\cmidrule{4-12}
&&
& \textcolor{black}{\boldxmark} \textcolor{black}{\boldxmark}               
& \textcolor{black}{\boldcheckmark} \textcolor{black}{\boldcheckmark}       
& \textcolor{black}{\boldcheckmark} \textcolor{black}{\boldcheckmark}       
& \textcolor{black}{\boldxmark} \textcolor{black}{\boldcheckmark}           
& \textcolor{black}{\boldxmark} \textcolor{black}{\boldcheckmark}           
& \textcolor{black}{\boldxmark} \textcolor{black}{\boldcheckmark}           
& \textcolor{black}{\boldxmark} \textcolor{black}{\boldxmark}               
& \textcolor{black}{\boldxmark} \textcolor{black}{\boldxmark}               
& \textcolor{black}{\boldxmark} \textcolor{black}{\boldxmark}               
\\
\midrule
\multirow{5}{*}{\makecell{Colored \\ MNIST}} 
& 0.5 & & 27.04 & 26.99 & 29.37 & 15.91 & 12.56 & 23.37 & 59.05 & 57.98 & \underline{\textbf{64.25}} \\ & 1.0 & & 44.11 & 40.74 & 53.13 & 33.29 & 23.06 & 41.66 & 74.88 & 72.14 & \underline{\textbf{76.54}} \\ & 2.0 & & 63.05 & 56.96 & 67.48 & 46.00 & 40.97 & 56.37 & 81.58 & 75.00 & \underline{\textbf{82.20}} \\ & 5.0 & & 79.48 & 76.20 & 79.41 & 77.29 & 70.25 & 77.19 & 85.44 & 82.46 & \underline{\textbf{88.45}} \\ & 20.0 & & \underline{\textbf{92.01}} & 89.82 & 91.41 & 90.07 & 90.80 & 88.51 & 89.95 & 91.13 & 90.43 \\  
\midrule
\multirow{5}{*}{\makecell{BFFHQ}} 
& 0.5 & & 48.78 & 53.32 & 51.62 & 52.84 & \textbf{54.52} & 53.36 & \underline{52.52} & 46.78 & 47.94 \\ & 1.0 & & 57.26 & 57.28 & 59.98 & 60.28 & 63.00 & 58.92 & 60.72 & 58.02 & \underline{\textbf{63.14}} \\ & 2.0 & & 64.32 & 69.18 & 65.90 & 67.78 & \textbf{73.56} & 68.46 & 67.48 & 66.06 & \underline{71.96} \\ & 5.0 & & \underline{80.32} & 82.60 & 78.24 & 81.54 & \textbf{84.98} & 82.52 & 69.38 & 78.88 & 73.36 \\ & 20.0 & & \underline{92.92} & 93.58 & 93.24 & 92.78 & \textbf{94.44} & 92.98 & 81.36 & 89.32 & 90.48 \\
\midrule
\multirow{5}{*}{\makecell{Waterbirds}} 
& 0.5 & & 37.18 & 38.39 & 40.72 & 54.18 & 42.63 & \textbf{77.69} & 56.00 & 56.39 & \underline{65.99} \\ & 1.0 & & 37.91 & 42.05 & 47.15 & 71.25 & 38.80 & \textbf{74.86} & 62.19 & 53.93 & \underline{64.74} \\ & 2.0 & & 47.48 & 41.76 & 74.86 & 65.50 & 44.96 & \textbf{76.43} & \underline{75.68} & 37.89 & 29.82 \\ & 5.0 & & 55.93 & 49.56 & 64.69 & 59.96 & 50.79 & \textbf{77.52} & 67.78 & \underline{71.02} & 52.85 \\ & 20.0 & & 69.47 & 71.24 & 70.71 & 71.00 & 68.92 & 67.97 & \underline{\textbf{72.04}} & 70.86 & 67.20 \\ 
\midrule
\multirow{3}{*}{\makecell{BAR}} 
& 1.0 & & 64.79 & - & - & 63.29 & 65.14 & 57.14 & 65.36 & \underline{\textbf{66.50}} & 66.07 \\ & 5.0 & & 82.71 & - & - & 78.86 & 79.36 & 74.86 & 82.36 & 83.79 & \underline{\textbf{84.43}} \\ & 20.0 & & 90.51 & - & - & 88.38 & 88.99 & 84.24 & \underline{\textbf{91.82}} & 91.62 & 91.01 \\
\midrule
\bottomrule
\end{tabular}
}
\caption{Image classification accuracy evaluated with only bias-conflicting samples in the unbiased tests sets. Additional details of this table are identical to those of Table~\ref{tab:main_f1_acc}.}
\vspace{-0.3cm}
\label{tab:main_f1_conflict}
\end{table*}

\begin{table*}[t!]
\centering
\scalebox{0.8}{
\begin{tabular}{cc|cc|cc|cc}
\toprule
\multicolumn{2}{c|}{\multirow{3}{*}{\makecell{Evaluation Metric}}} & \multicolumn{4}{c|}{Accuracy} & \multicolumn{2}{c}{\multirow{3}{*}{\makecell{Equality of Opportunity}}} \\
\cmidrule{3-6}
& & \multicolumn{2}{c|}{Unbiased}  & \multicolumn{2}{c|}{Bias-conflicting} & & \\
\cmidrule{1-8}
\multicolumn{2}{c|}{Tuning Criterion}
& Avg. Acc.
& AC score
& Avg. Acc.
& AC score
& Avg. Acc.
& AC score
\\
\midrule
\multirow{5}{*}{\makecell{Colored \\ MNIST}} 
& 0.5 pct & 32.04 & \textbf{41.34} & 24.31 & \textbf{35.17} & 0.2481 & \textbf{0.3811} \\ 
& 1.0 pct & 45.61 & \textbf{55.61} & 39.53 & \textbf{51.06} & 0.4058 & \textbf{0.5457} \\ 
& 2.0 pct & 62.66 & \textbf{66.66} & 58.63 & \textbf{63.29} & 0.6041 & \textbf{0.6610} \\ 
& 5.0 pct & 81.27 & \textbf{81.46} & 79.35 & \textbf{79.58} & 0.8088 & \textbf{0.8128} \\ 
& 20.0 pct & 91.28 & \textbf{91.34} & 90.40 & \textbf{90.46} & 0.9135 & \textbf{0.9141} \\
\midrule
\multirow{5}{*}{\makecell{BFFHQ}} 
& 0.5 pct & \textbf{74.35} & 74.25 & 51.12 & \textbf{51.30} & 0.5234 & \textbf{0.5269} \\ 
& 1.0 pct & 78.08 & \textbf{78.18} & 58.30 & \textbf{59.84} & 0.5958 & \textbf{0.6211} \\ 
& 2.0 pct & 81.72 & \textbf{81.96} & 66.13 & \textbf{68.30} & 0.6794 & \textbf{0.7166} \\ 
& 5.0 pct & 86.43 & \textbf{87.13} & 76.16 & \textbf{79.09} & 0.7857 & \textbf{0.8307} \\ 
& 20.0 pct & \textbf{93.14} & \textbf{93.14} & 90.65 & \textbf{91.23} & 0.9437 & \textbf{0.9528} \\
\midrule
\multirow{5}{*}{\makecell{Waterbirds}} 
& 0.5 pct & 60.77 & \textbf{66.40} & 33.31 & \textbf{52.13} & 0.3779 & \textbf{0.6006} \\ 
& 1.0 pct & 62.41 & \textbf{67.53} & 36.07 & \textbf{54.76} & 0.4100 & \textbf{0.6263} \\ 
& 2.0 pct & 61.57 & \textbf{68.19} & 34.43 & \textbf{54.93} & 0.3733 & \textbf{0.5783} \\
& 5.0 pct & 66.34 & \textbf{72.57} & 42.78 & \textbf{61.12} & 0.4803 & \textbf{0.6688} \\ 
& 20.0 pct & 76.63 & \textbf{78.13} & 65.13 & \textbf{69.93} & 0.7163 & \textbf{0.7834} \\
\midrule
\bottomrule
\end{tabular}
}
\vspace{-0.2cm}
\caption{Comparisons on accuracy and fairness between selecting model checkpoints and hyper-parameters based on the average validation set accuracy (denoted as Avg. Acc.) and AC score. We report the average of accuracy and fairness of 9 debiasing methods.}
\vspace{-0.3cm}
\label{tab:main_difference}
\end{table*}

\subsection{Experimental settings}
\label{sec:experimental_setting}
We utilize a 3-layer multi-layer perceptron (MLP) for Colored MNIST and ResNet18 for the rest of the datasets.
Due to an excessively small number of images in the training set, we use an Imagenet-pretrained ResNet18 for BAR while using a randomly initialized one for BFFHQ and Waterbirds.
For each dataset and each bias severity, we select hyper-parameters and model checkpoints based on AC score using the validation set. 
We conducted five independent trials and reported the mean of the five trials for each experiment. 
For the BAR dataset, we could not train debiasing methods which require bias labels during the training phase (denoted as `-') since the BAR dataset does not have bias annotations in the training set.
In our Supplementary, we include results with the standard deviation, which we intentionally omit due to the space limit.
We also include further implementation details of our experimental setting in our Supplementary.

\section{Discussion and Analysis}


\subsection{Utilizing AC score for Tuning Criterion}
\label{sec:ac_f1_score_tuning}
Table~\ref{tab:main_f1_acc} and Table~\ref{tab:main_f1_conflict} show the accuracy of the unbiased test set and that of bias-conflicting samples in the test set, respectively, both tuned by AC score using the validation set. 
In each table, the cross and check marks indicate whether a debiasing method 1) requires explicit bias labels and 2) predefines a bias type (\textit{e.g.,} color bias) during the training stage. 
The bold digits indicate the best result among all debiasing methods. 
The underlined digits refer to the best result between debiasing methods that do not utilize bias labels or predefine bias types.

The justifications for utilizing AC score for the tuning criterion are as follows.
First, it does not require preprocessing the validation set to have the equal number of bias-aligned samples and bias-conflicting samples.
In order to construct such a validation set, we need to know whether a given training sample is a bias-aligned one or a bias-conflicting one which requires a non-trivial amount of labor. 
Instead, when using AC score, we can randomly sample training samples and construct the validation set.
While our method still requires a small amount of labor for annotating whether a sample in the validation set is a bias-aligned one or not, we discuss such limitation in Section~\ref{limitation}. 
Second, using AC score improves the debiasing performance and the fairness metric compared to using the average accuracy of the validation set for the tuning criterion.
While the average accuracy of the validation set mainly reflects the performance of the bias-aligned samples, AC score reflects the performance of both bias-aligned samples and bias-conflicting samples.

Table~\ref{tab:main_difference} shows 1) the accuracy of the unbiased test set, 2) that of only bias-conflicting samples in the unbiased test set, and 3) the equality of opportunity, one of the widely used fairness metrics~\cite{fair-eo, fair-train}. 
We compare the results tuned with the average accuracy (denoted as Avg. Acc.) and AC score using the validation set. 
Equality of opportunity (E.O) is formulated as 

\begin{equation}
    \text{E.O} = \frac{P(\hat{Y}=1|B=0, Y=1)}{P(\hat{Y}=1|B=1, Y=1)},
\end{equation}
where $Y$ indicates the target class and $B$ indicates the bias attribute. 
A classifier makes fair predictions with high equality of opportunity.

We averaged the evaluation metrics (\textit{i.e.,} accuracy and E.O) of 9 debiasing methods for each tuning criterion (\textit{i.e.,} Avg. Acc. and AC score).
We observe that using AC score as the tuning criterion generally improves the three evaluation metrics regardless of the bias severity. 
Since using AC score as the tuning criterion does not require preprocessing the validation set and improves the debiasing performance and the fairness metric, we highly encourage future studies to utilize it for the tuning criterion.

\subsection{Performance Comparisons}
\label{sec:performance_compare}
Following are several findings we observed from the extensive experiments in our study. 
First, there is no debiasing method which outperforms other baseline methods over all datasets and varying bias severities. 
A given debiasing method may show superior performance on a certain dataset while showing degraded performance on other datasets. 
This indicates that there exists a room for future debiasing researchers to propose debiasing methods to outperform other methods regardless of bias types and bias severities. 
Second, recent debiasing methods which follow the assumption that bias labels and predefined bias types are unknown during the training phase fail to show superior performance in BFFHQ and Waterbirds compared to the ones which use bias labels or predefine bias types.
Since the recent studies follow this assumption, we believe that future researchers may improve on datasets that the recent debiasing studies under such an assumption fail to show superior performance.  

\begin{table*}[t!]
\centering
\scalebox{0.85}{
\begin{tabular}{c|ccccc|ccccc}
\toprule
\multirow{2.5}{*}{Architecture} 
& \multicolumn{5}{c|}{Acc. Diff. $\downarrow$} 
& \multicolumn{5}{c}{Equality of Opportunity $\uparrow$} 
\\
\cmidrule{2-6}
\cmidrule{7-11}
& 0.5\%
& 1.0\%
& 2.0\%
& 5.0\%
& 20.0\%
& 0.5\%
& 1.0\%
& 2.0\%
& 5.0\%
& 20.0\%
\\
\midrule
Colored MNIST         & 72.94 & 55.76 & 36.77 & 20.12 & 7.28  & 0.27 & 0.44 & 0.63 & 0.80 & 0.93 \\ 
BFFHQ                 & 49.36 & 41.58 & 34.30 & 18.20 & 5.10  & 0.50 & 0.58 & 0.65 & 0.82 & 0.95 \\ 
WaterBirds            & 53.99 & 52.73 & 42.76 & 34.62 & 19.53 & 0.38 & 0.39 & 0.48 & 0.58 & 0.75 \\ 
BAR                   & -     & 33.14 &   -   & 16.29 & 9.09  &   -  & 0.66 &   -  & 0.84 & 0.91 \\ 
\midrule
\bottomrule
\end{tabular}
}
\vspace{-0.2cm}
\caption{Performances of vanilla method depending on the bias severity. We report 1) Acc. Diff. which indicates the difference of accuracy between the bias-aligned samples and the bias-conflicting samples and 2) the equality of opportunity. We observe that there still exists room to be improved under the ratio of 20\% of bias-conflicting samples.}
\label{tab:main_severity_diff}
\end{table*}

\begin{table*}[t!]
\centering
\scalebox{0.75}{
\begin{tabular}{cc|cccccc|ccccc}
\toprule
\multirow{2.5}{*}{Dataset} 
& \multirow{2.5}{*}{Architecture} 
& \multicolumn{6}{c|}{Kruskal-Wallis statistics $\uparrow$} 
& \multicolumn{5}{c}{$p$ value $\downarrow$} 
\\
\cmidrule{4-8}
\cmidrule{9-13}
&&
& 0.5\%
& 1.0\%
& 2.0\%
& 5.0\%
& 20.0\%
& 0.5\%
& 1.0\%
& 2.0\%
& 5.0\%
& 20.0\%
\\
\midrule
\multirow{2}{*}{\makecell{Colored \\ MNIST}} 
& MLP         & & 39.42 & \textbf{39.55} & \textbf{41.71} & \textbf{39.52} & 30.78 & 4.12e-06 & \textbf{3.89e-06} & \textbf{1.54e-06} & \textbf{3.93e-06} & 1.54e-04 \\ 
& Simple Conv & & \textbf{41.42} & 37.79 & 40.15 & 33.80 & \textbf{33.56} & \textbf{1.74e-06} & 8.23e-06 & 3.01e-06 & 4.41e-05 & \textbf{4.88e-05} \\ 
\midrule
\bottomrule
\end{tabular}
}
\vspace{-0.2cm}
\caption{Comparisons between using MLP and a stack of four convolutional layers (denoted as `Simple Conv'). The higher the Kruskal-Wallis statistics and the lower P values are, we obtain bigger performance differences between each debiasing method. The best results are marked in bold. We observe that utilizing MLP obtains bigger difference between each method in three out of five ratios.}
\vspace{-0.45cm}
\label{tab:main_cmnist}
\end{table*}

\subsection{Further Improvements on Low Bias Severity}
\label{sec:low_severity}
Table~\ref{tab:main_f1_acc} and Table~\ref{tab:main_f1_conflict} also show that the recent debiasing methods, those which do not utilize bias labels or predefine bias types, fail to consistently outperform the vanilla method on datasets with low bias severity (20\% of bias-conflicting samples).
To be more specific, they fail to outperform the vanilla method with the unbiased test set accuracy in Colored MNIST, BFFHQ, and Waterbirds (Table~\ref{tab:main_f1_acc}). 
They even achieve lower accuracy than the vanilla method when evaluated with only bias-conflicting samples in Colored MNIST and BFFHQ (Table~\ref{tab:main_f1_conflict}). 
This is especially surprising considering the fact that reweighting-based algorithms~\cite{nam2020learning, disentangled, BiasEnsemble} emphasize the bias-conflicting samples during the training phase.
Even without such emphasis, utilizing the vanilla method shows superior performance on bias-conflicting samples in Colored MNIST and BFFHQ when the bias severity is low. 

One might question whether a dataset with the ratio of 20\% is not a biased dataset, and it is inappropriate to use it for the evaluation.
Table~\ref{tab:main_severity_diff} demonstrates that the bias severity with the ratio of 20\% is still biased.
For each bias severity, we computed 1) the difference between the accuracy of the bias-aligned samples and that of the bias-conflicting samples (denoted as Acc. Diff.) and 2) the equality of opportunity using the vanilla method.
Ideally, unbiased datasets achieve 0 and 1 for the Acc. Diff and equality of opportunity, respectively.
We observe that the Acc. Diff and the equality of opportunity for datasets with the ratio of 20\% need to be improved. 
Therefore, we believe debiasing under low bias severity would be one of the next topics to be explored in debiasing. 

\begin{table*}[h]
\centering
\scalebox{0.85}{
\begin{tabular}{ccccccc}
\toprule

& $f_B$ Arch.
& $f_B$ Valid B.A
& $f_B$ Valid B.C
& $f_B$ Acc. Diff.
& $f_D$ Test Unbiased.
& $f_D$ Test B.C
\\
\midrule
& MLP  & 96.28\stdv{0.10} & 57.00\stdv{2.74} & \textbf{39.28} & 80.47\stdv{1.33} & \textbf{66.56}\stdv{2.55} \\ 
& Simple Conv & 95.48\stdv{0.15} & 58.00\stdv{2.74} & 37.48 & \textbf{80.75}\stdv{1.70} & 64.44\stdv{2.93} \\ 
\midrule
& ResNet18  & 99.45\stdv{0.07} & 74.00\stdv{4.18} & 25.45 & 73.37\stdv{3.19} & 60.72\stdv{2.20} \\ 
\midrule
\bottomrule
\end{tabular}
}
\vspace{-0.15cm}
\caption{Metrics for bias amplification and debiasing performance. $f_B$ and $f_D$ indicate the biased model and the debiased model used in LfF~\cite{nam2020learning}. B.A and B.C are the abbreviations of bias-aligned samples and bias-conflicting samples, respectively. We observe that simply utilizing small backbones for $f_B$ improves the debiasing performance of $f_D$.}
\vspace{-0.3cm}
\label{tab:main_fb}
\end{table*}

\subsection{Comparisons between architectures for Colored MNIST}
\label{sec:cmnist_mlp_conv}
While Colored MNIST has been widely utilized as a small dataset for evaluating a debiasing method, previous studies do not share a common neural network architecture.
To be more specific, several studies~\cite{nam2020learning, disentangled, BiasEnsemble} utilize a 3-layer MLP while others use a stack of 4 Convolutional layers~\cite{bahng2019rebias, EnD, softcon}.
This section finds out which neural network architecture is more adequate to utilize for Colored MNIST. 

We conducted experiments on Colored MNIST with both MLP and a stack of Convolutional layers using AC score.
Then, for each neural network architecture, we conducted the Kruskal-Wallis test~\cite{kruskal-wallis} using five independent trials of each debiasing algorithm in order to find out whether there is a statistical difference between the average accuracy of 9 debiasing methods. 
We conducted the Kruskal-Wallis test instead of One-way Anova~\cite{one-way-anova} since each group (\textit{i.e.,} debiasing method) only contains five trials, indicating that it is hard to assume that the values follow the normal distribution.
Note that we obtain a bigger difference between each method as we obtain bigger Kruskal-Wallis statistics (\textit{i.e.,} lower $p$ value).

Table~\ref{tab:main_cmnist} shows that utilizing MLP obtains bigger Kruskal-Wallis statistics compared to using a stack of Convolutional layers (denoted as `Simple Conv') in 3 out of 5 ratios. 
Since we can obtain bigger differences between each method with MLP compared to Convolutional layers, we utilize MLP as the neural network architecture for Colored MNIST. 
Obtaining large performance differences between methods enables researchers to confirm the effectiveness of their proposed method effortlessly during research. 
Additionally, training with MLP reduces a significant amount of training time compared to using Convolutional layers due to its few number of learnable parameters which is beneficial for debiasing researchers when proposing a new method.  
However, since both architectures achieve $p$ values under 0.05~\footnote{In null-hypothesis significance testing, achieving $p$ values under 0.05 indicates to reject the null hypothesis. In our case, it indicates there exists a difference among accuracies of debiasing methods.} for all ratios, we want to clarify that both neural network architectures are adequate for the evaluation with Colored MNIST. 

\section{Future Direction and Limitation}
\subsection{Utilizing Small Backbones for $f_B$}
\label{sec:fb_small}
Along with the findings based on our extensive experiments, we also provide future research directions in debiasing.
Reweighting-based debiasing algorithms~\cite{nam2020learning, disentangled, BiasEnsemble} utilize a biased classifier $f_B$ to upweight (\textit{i.e.,} imposing high weights on loss values) bias-conflicting samples when training a debiased classifier $f_D$.
With such a training scheme, Lee~\textit{et al.} emphasized the importance of amplifying bias for $f_B$ in order to improve the debiasing performance of $f_D$~\cite{BiasEnsemble}.
In the meanwhile, several studies point out that networks which utilize small backbones are vulneBiasEnsemblele to being overfitted to bias or to relying on the shortcut learning when making predictions~\cite{learning_other_mistakes, towards-nlu-bias}. 

Inspired by the previous studies, we explored whether reweighting-based debiasing algorithms can improve debiasing performance of $f_D$ by using small architectures for $f_B$.
We used BFFHQ for the experiment.
By fixing the architecture of $f_D$ as ResNet18, we conducted experiments with LfF~\cite{nam2020learning} by modifying the architecture of $f_B$ from ResNet18 to 3-layer MLP and a stack of 4 Convolutional layers. 
As mentioned in the work of Lee~\textit{et al.}~\cite{BiasEnsemble}, $f_B$ is likely to be overfitted to the bias as it achieves 1) high classification accuracy on bias-aligned samples and 2) low classification accuracy on bias-conflicting ones. 
$f_B$ Valid B.A and $f_B$ Valid B.C in Table~\ref{tab:main_fb} indicate the accuracy of bias-aligned samples and that of bias-conflicting samples in the validation set, respectively. 
Acc. Diff. in Table~\ref{tab:main_fb} indicates the difference between the two values, so $f_B$ is likely to be overfitted to the bias as Acc. Diff. increases. 
Table~\ref{tab:main_fb} shows that \emph{simply} utilizing small neural network architectures for $f_B$ (\textit{e.g.,} a 3-layer MLP or a stack of 4 convolutional layers) improves the debiasing performance of $f_D$ compared to using ResNet18, the original architecture, for $f_B$.
The main reason is due to the high $f_B$ Diff., indicating that using small backbones enables to further encourage $f_B$ to be overfitted to the bias. 
As an additional benefit, utilizing smaller backbones requires 1) fewer number of learnable parameters and 2) less training time compared to large architectures. 
We believe that such a valuable finding provides directions for future debiasing researchers when deciding the architecture of $f_B$ in reweighting-based debiasing algorithms.



\subsection{Annotations on Bias-aligned/conflicting Samples in Validation Set}
\label{limitation}
One of the limitations of our work is that we require annotations to identify whether a sample in the validation set is a bias-aligned sample or a bias-conflicting sample.
Note that previous work~\cite{EnD} which constructs the validation set to have the equal number of bias-aligned samples and bias-conflicting samples need to have such annotations on \emph{all training} samples.
However, utilizing AC score for the tuning criterion only requires such annotations on the validation set which is a small subset of the training set.
To be more specific, we randomly sample data instances to construct the validation set.
Then, we assign annotations on whether a given sample in the validation set is a bias-aligned one or the bias-conflicting one. 
Assigning such annotations on the validation set instead of the training set saves a significant amount of labor. 
To the best of our knowledge, this is the first work in debiasing to take account of both bias-aligned samples and bias-conflicting ones when selecting the tuning criterion without adjusting the number of samples. 
We believe that our work inspires future researchers to propose tuning criterion adequate for debiasing even without assigning such annotations on the validation set. 

\section{Conclusion}
In this work, we proposed experimental settings which improve the evaluation of debiasing in image classification.
First, we propose `Align-Conflict (AC) score' which considers both accuracy of bias-aligned samples and bias-conflicting samples in the validation set for selecting hyper-parameters and model checkpoints.
Second, since real-world datasets may include a non-trivial number of bias-conflicting samples and the current debiasing methods fail to consistently outperform the vanilla method with the low bias severity, we include evaluations with the low bias severity and highly encourage the future debiasing studies to focus on improving debiasing performance with such a bias severity.  
Third, we standardized the inconsistent experimental settings found in the previous studies such as datasets and neural network architectures. 
We believe that the valuable findings and lessons obtained through our comprehensive study inspire future debiasing researchers to further push the state-of-the-art performances in debiasing.


\clearpage
{\small
\bibliographystyle{ieee_fullname}
\bibliography{egbib}
}

\clearpage
\appendix
This supplementary presents the details of the debiasing methods, datasets used in our work, implementation details, and results of Table~2 and Table~3 of the main paper with the standard deviation. 

\section{Further Details on Debiasing Methods}
We briefly summarize the debiasing methods used in our work. We note a biased classifier and a debiasied classifier as $f_B$ and $f_D$, respectively. 

\noindent \textbf{LNL~\cite{LNL}} minimizes the mutual information between the feature embeddings of intrinsic attributes and those of bias attributes by using an additional branch of classifier which learns bias attributes. \\
\noindent \textbf{EnD~\cite{EnD}} proposes a regularizer which disentangles the representation space into features learning intrinsic attributes and bias attributes. \\
\noindent \textbf{HEX~\cite{wang2018hex}} utilizes a reverse gradient method and projects the features of intrinsic attributes orthogonal to those of bias attributes. \\
\noindent \textbf{ReBias~\cite{bahng2019rebias}} trains $f_D$ to be independent from $f_B$, a model with small receptive fields to focus on color and texture, by using Hilbert-Schmidt Independence Criterion~\cite{hsic}. \\ 
\noindent \textbf{Softcon~\cite{softcon}} trains $f_D$ by emphasizing the bias-conflicting samples based on the cosine distance between a pair of samples at the feature space of a pretrained $f_B$. \\ 
\noindent \textbf{LfF~\cite{nam2020learning}} trains $f_D$ by up-weighting the losses of the bias-conflicting samples and down-weighting those of the bias-aligned samples based on the intuition that bias is easy to learn. \\  
\noindent \textbf{DisEnt~\cite{disentangled}} proposes a feature-level augmentation of bias-conflicting samples based on the observation that diversity of bias-conflicting samples is crucial in debiasing. \\  
\noindent \textbf{BiasEnsemble~\cite{BiasEnsemble}} emphasizes the importance of amplifying bias for $f_B$ in order to improve the debiasing performance of $f_D$, an intuition applicable to existing reweighting-based debiasing algorithms.  \\ 

\section{Further Details on Datasets}
\begin{figure*}[t]
    \centering
    \includegraphics[width=\textwidth]{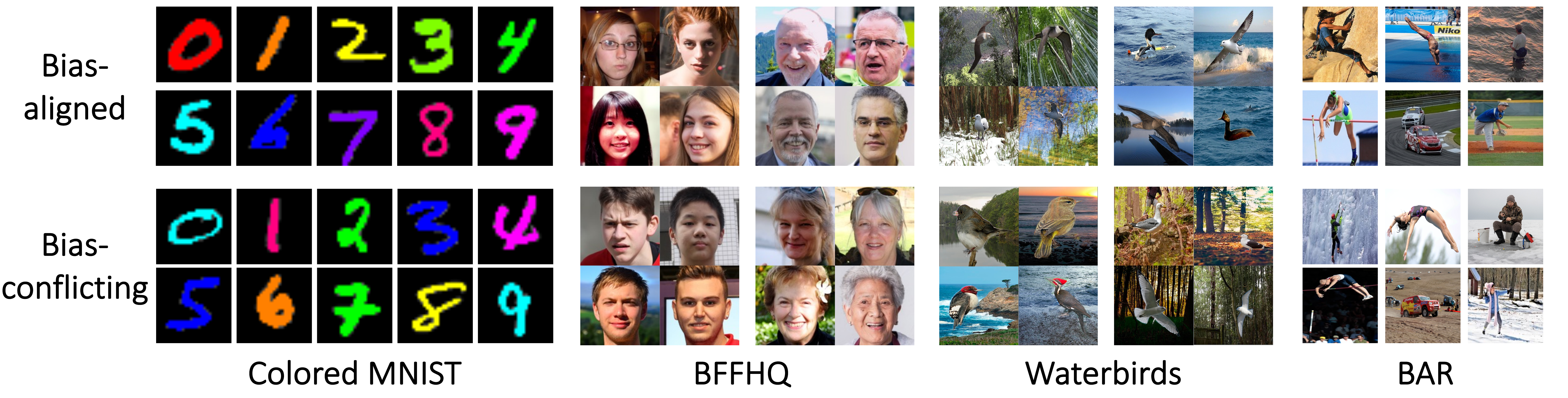}
    \caption{Datasets used in our work. Each grid of Colored MNIST and BAR indicate each class while each two columns indicate each class for BFFHQ and Waterbirds.}
    \label{fig:supple_dataset}
\end{figure*}

We describe the datasets used in our work. 
Fig.~\ref{fig:supple_dataset} shows the images of the datasets. 
Also, Table~\ref{tab:supple_dataset} shows the number of bias-aligned samples and bias-conflicting samples corresponding to each bias severity.

\noindent \textbf{Colored MNIST} has digits and colors for the intrinsic and bias attributes, respectively.
A certain color is highly correlated with a certain digit (\textit{e.g.,} red color frequently appearing in images of 0).
For building the Colored MNIST, we follow the dataset used in Lee~\textit{et al.}~\cite{disentangled}. 

\noindent \textbf{BFFHQ} has age (i.e., young and old) and gender for the intrinsic and bias attributes, respectively. Old male and young female correspond to the bias-aligned samples while young male and old female are the bias-conflicting samples.
BFFHQ is modified from FFHQ~\cite{stylegan}. 
We follow the dataset used in Kim~\textit{et al.}~\cite{biaswap} for biased FFHQ. 

\begin{table*}[t!]
\centering
\scalebox{0.85}{
\begin{tabular}{cc|ccccc|ccccc|c}
\toprule
\multirow{2.5}{*}{Dataset} 
& \multirow{2.5}{*}{Data samples} 
& \multicolumn{5}{c|}{Training set} 
& \multicolumn{5}{c|}{Validation set} 
& Test set
\\
\cmidrule{3-13}
&
& 0.5\%
& 1.0\%
& 2.0\%
& 5.0\%
& 20.0\%
& 0.5\%
& 1.0\%
& 2.0\%
& 5.0\%
& 20.0\%
& Unbiased
\\
\midrule
\multirow{2}{*}{\makecell{Colored \\ MNIST}} 
& Bias-aligned         & 44,770 & 44,500 & 44,100 & 42,750 & 36,000 & 4,970 & 4,950 & 4,900 & 4,750 & 4,000 & 1,037 \\ 
& Bias-conflicting     & 230    & 450    & 900    & 2,250  & 9,000  & 30    & 50    & 100   & 250   & 1,000 & 8,963 \\ 
\midrule
\multirow{2}{*}{\makecell{BFFHQ}} 
& Bias-aligned         & 17,910 & 17,820 & 17,640 & 17,100 & 14,400 & 1,990 & 1,980 & 1,960 & 1,900 & 1,600 & 1,000 \\ 
& Bias-conflicting     & 90     & 180    & 360    & 900    & 3,600  & 10    & 20    & 40   & 100    & 400   & 1,000 \\ 
\midrule
\multirow{2}{*}{\makecell{Waterbirds}} 
& Bias-aligned         & 4,771 & 4,747 & 4,699 & 4,555 & 3,836 & 1,193 & 1,187 & 1,175 & 1,139 & 959 & 4,327 \\ 
& Bias-conflicting     & 24    & 48    & 96    & 240  & 959  & 6    & 12    & 24   & 60   & 240 & 4,327 \\ 
\midrule
\multirow{2}{*}{\makecell{BAR}} 
& Bias-aligned         & - & 1,493 & - & 1,493 & 1,493 & - & 168 & - & 168 & 168 & 280 \\ 
& Bias-conflicting     & - & 13    & - & 76  & 370  & - & 6    & - & 7   & 41 & 280 \\ 
\midrule

\bottomrule
\end{tabular}
}
\caption{The number of bias-aligned samples and bias-conflicting samples corresponding to each dataset and bias severity.}
\label{tab:supple_dataset}
\end{table*}

\noindent \textbf{Waterbirds} has the bird type (\textit{i.e.,} waterbird and landbird) and background (\textit{i.e.,} the water background and the land background) for the intrinsic and bias attributes, respectively. 
Following the previous work~\cite{sagawa2019distributionally}, the bird images are sampled from CUB dataset~\cite{cub} and the background images are sampled from Places dataset~\cite{places}. 
Then, bird images are synthesized with the background images.
Waterbirds in the water background and landbirds in the land background are the bias-aligned samples while the waterbirds in the land background and the landbirds in the water background are the bias-conflicting samples. 

\noindent \textbf{BAR} has the action and background for the intrinsic and bias attributes, respectively. Nam~\textit{et al.}~\cite{nam2020learning} did not evaluate debiasing methods according to different ratios of bias-conflicting samples in BAR. 
We intentionally preprocessed BAR dataset with different ratios of bias-conflicting samples for the unified experimental setting as done for other datasets. 

\newpage

\section{Further Implementation Details}
We describe further implementation details of our work. 
For Colored MNIST, we use the batch size of 256, image size of $28\times28$, 200 training epochs, and did not apply data augmentation methods. 
For the rest of the datasets, we use the batch size of 64, image size of $224\times224$, 100 training epochs, and apply random crop and random horizontal flip for the data augmentation methods. 
Regarding the image processing, we normalized each channel with the mean of (0.485, 0.456, 0.406) and the standard deviation of (0.229, 0.224, 0.225). 
We train the models using the Adam~\cite{adam} optimizer with the default parameters provided in the Pytorch library. 
Since we empirically confirm that learning rate significantly determines the debiasing performance, we first select the learning rate which achieves the best AC score with the validation set. 
With the selected learning rate, we then tune the rest of the hyper-parameters mentioned in the main paper of each debiasing method using the AC score of the validation set.

\section{Results with Standard Deviations}
Due to the space limit, we intentionally omitted the standard deviations in the main paper. Table~\ref{supple:main_f1_acc} and Table~\ref{supple:main_f1_acc_conflict} reports the results of Table~2 and Table~3 of the main paper, respectively, with the standard deviations included. 

\begin{sidewaystable}[]
\vspace{9.5cm}
\centering
\scalebox{1.0}{
\begin{tabular}{ccccccccccccc}
\toprule
\multirow{2.5}{*}{Dataset} 
& \multirow{2.5}{*}{Ratio (\%)} 
&
& Vanilla
& LNL~\cite{LNL}
& EnD~\cite{EnD}
& HEX~\cite{wang2018hex}
& ReBias~\cite{bahng2019rebias}
& Softcon~\cite{softcon}
& LfF~\cite{nam2020learning}
& DisEnt~\cite{disentangled}
& BiasEnsemble~\cite{BiasEnsemble}

\\
\cmidrule{4-12}
&&
& \textcolor{black}{\boldxmark} \textcolor{black}{\boldxmark}               
& \textcolor{black}{\boldcheckmark} \textcolor{black}{\boldcheckmark}       
& \textcolor{black}{\boldcheckmark} \textcolor{black}{\boldcheckmark}       
& \textcolor{black}{\boldxmark} \textcolor{black}{\boldcheckmark}           
& \textcolor{black}{\boldxmark} \textcolor{black}{\boldcheckmark}           
& \textcolor{black}{\boldxmark} \textcolor{black}{\boldcheckmark}           
& \textcolor{black}{\boldxmark} \textcolor{black}{\boldxmark}               
& \textcolor{black}{\boldxmark} \textcolor{black}{\boldxmark}               
& \textcolor{black}{\boldxmark} \textcolor{black}{\boldxmark}               
\\
\midrule
\multirow{5}{*}{\makecell{Colored \\ MNIST}} 
& 0.5 pct &  & 34.61\stdv{2.79} & 34.54\stdv{4.78} & 36.69\stdv{3.12} & 24.58\stdv{4.17} & 21.62\stdv{2.55} & 31.29\stdv{1.69} & 62.01\stdv{6.76} & 59.99\stdv{1.71}& \underline{\textbf{66.71}}\stdv{2.66}\\ & 1.0 pct &  & 49.87\stdv{0.73} & 46.83\stdv{4.01} & 57.87\stdv{1.94} & 39.80\stdv{12.63} & 31.01\stdv{4.45} & 47.66\stdv{4.50} & 76.17\stdv{1.01} & 73.30\stdv{2.66}& \underline{\textbf{77.94}}\stdv{1.39}\\ & 2.0 pct &  & 66.84\stdv{1.34} & 61.35\stdv{3.10} & 70.71\stdv{0.69} & 50.76\stdv{13.65} & 47.04\stdv{5.78} & 60.79\stdv{2.01} & 82.63\stdv{1.05} & 76.86\stdv{3.59}& \underline{\textbf{82.98}}\stdv{0.66}\\ & 5.0 pct &  & 81.56\stdv{0.73} & 78.56\stdv{1.04} & 81.50\stdv{1.82} & 79.58\stdv{2.63} & 73.24\stdv{2.44} & 79.46\stdv{0.73} & 86.43\stdv{1.38} & 84.05\stdv{0.54}& \underline{\textbf{88.74}}\stdv{0.44}\\ & 20.0 pct & & \underline{\textbf{92.77}}\stdv{0.30} & 90.73\stdv{0.63} & 92.22\stdv{0.82} & 91.02\stdv{0.22} & 91.68\stdv{0.56} & 89.57\stdv{0.39} & 90.76\stdv{0.69} & 91.97\stdv{0.33} & 91.32\stdv{1.58}\\ 
\midrule
\multirow{5}{*}{\makecell{BFFHQ}} 
& 0.5 pct & & \underline{73.46}\stdv{0.66} & 76.20\stdv{0.70} & 75.41\stdv{1.03} & 76.09\stdv{0.80}& \textbf{76.64}\stdv{2.49} & 76.09\stdv{0.88} & 73.07\stdv{1.96} & 71.72\stdv{1.39} & 69.61\stdv{2.55}\\ & 1.0 pct &  & 78.05\stdv{0.64} & 78.16\stdv{0.95} & 79.51\stdv{0.79} & 79.69\stdv{1.04}& \textbf{80.98}\stdv{1.35} & 78.75\stdv{0.70} & 73.37\stdv{3.57} & 76.10\stdv{1.48}& \underline{78.98}\stdv{1.07}\\ & 2.0 pct & & \underline{81.47}\stdv{0.43} & 83.93\stdv{1.21} & 82.42\stdv{1.33} & 83.40\stdv{0.88}& \textbf{86.31}\stdv{0.65} & 83.55\stdv{1.35} & 77.22\stdv{1.85} & 79.24\stdv{2.85} & 80.09\stdv{2.87}\\ & 5.0 pct & & \underline{89.42}\stdv{1.17} & 90.38\stdv{0.61} & 88.37\stdv{1.01} & 90.00\stdv{0.65}& \textbf{91.91}\stdv{0.30} & 90.65\stdv{0.98} & 78.55\stdv{2.04} & 85.10\stdv{2.63} & 79.80\stdv{2.06}\\ & 20.0 pct & & \underline{95.47}\stdv{0.40} & 95.75\stdv{0.28} & 95.66\stdv{0.35} & 95.22\stdv{0.36}& \textbf{96.43}\stdv{0.17} & 95.47\stdv{0.16} & 83.59\stdv{3.82} & 91.04\stdv{2.26} & 89.60\stdv{0.51}\\ 
\midrule
\multirow{5}{*}{\makecell{Waterbirds}} 
& 0.5 pct &  & 64.18\stdv{5.66} & 65.22\stdv{1.42} & 65.52\stdv{4.65} & 70.09\stdv{5.15} & 66.57\stdv{3.30}& \textbf{77.96}\stdv{0.26} & 59.80\stdv{4.32}& \underline{66.55}\stdv{5.14} & 61.73\stdv{8.91}\\ & 1.0 pct &  & 64.28\stdv{4.65} & 66.69\stdv{4.40} & 68.37\stdv{5.42} & 76.31\stdv{2.35} & 65.16\stdv{0.84}& \textbf{77.72}\stdv{0.71} & 61.69\stdv{2.95}& \underline{66.21}\stdv{4.15} & 61.34\stdv{5.29}\\ & 2.0 pct & & \underline{68.87}\stdv{2.80} & 67.29\stdv{2.21} & 77.02\stdv{2.33} & 74.08\stdv{5.71} & 67.88\stdv{1.01}& \textbf{78.03}\stdv{0.15} & 66.08\stdv{0.75} & 53.86\stdv{12.68} & 60.64\stdv{1.62}\\ & 5.0 pct &  & 73.24\stdv{2.66} & 70.66\stdv{2.06} & 74.87\stdv{3.22} & 74.40\stdv{3.10} & 70.95\stdv{1.64}& \textbf{77.97}\stdv{0.28} & 65.84\stdv{5.74}& \underline{74.35}\stdv{2.15} & 70.84\stdv{2.30}\\ & 20.0 pct & & \underline{79.23}\stdv{0.81}& \textbf{79.74}\stdv{0.65} & 79.54\stdv{0.75} & 79.40\stdv{0.64} & 79.30\stdv{0.72} & 77.90\stdv{1.42} & 72.94\stdv{3.71} & 77.22\stdv{2.41} & 77.95\stdv{1.21}\\
\midrule
\multirow{3}{*}{\makecell{BAR}} 
& 1.0 pct &  & 81.36\stdv{1.98} & - & - & 80.64\stdv{2.67} & 81.89\stdv{1.19} & 77.14\stdv{3.38} & 82.11\stdv{1.56}& \underline{\textbf{82.82}}\stdv{1.28} & 82.46\stdv{0.65}\\ & 5.0 pct &  & 90.86\stdv{0.50} & - & - & 88.39\stdv{1.73} & 89.07\stdv{0.87} & 86.50\stdv{0.83} & 90.71\stdv{0.69} & 91.25\stdv{0.99}& \underline{\textbf{91.46}}\stdv{0.50}\\ & 20.0 pct &  & 95.05\stdv{0.52} & - & - & 93.38\stdv{0.65} & 93.59\stdv{0.34} & 90.35\stdv{1.20}& \underline{\textbf{95.61}}\stdv{0.94} & 95.61\stdv{0.68} & 95.05\stdv{0.38}\\  
\midrule
\bottomrule
\end{tabular}
}
\vspace{-0.2cm}
\caption{Image classification accuracy evaluated on unbiased tests sets. The \textit{cross} and \textit{check} marks denote whether each algorithm requires 1) bias labels during training and 2) prior knowledge on bias types.}
\label{supple:main_f1_acc}
\end{sidewaystable}

\clearpage
\begin{sidewaystable}[]
\vspace{9.5cm}
\centering
\scalebox{1.0}{
\begin{tabular}{ccccccccccccc}
\toprule
\multirow{2.5}{*}{Dataset} 
& \multirow{2.5}{*}{Ratio (\%)} 
&
& Vanilla
& LNL~\cite{LNL}
& EnD~\cite{EnD}
& HEX~\cite{wang2018hex}
& ReBias~\cite{bahng2019rebias}
& Softcon~\cite{softcon}
& LfF~\cite{nam2020learning}
& DisEnt~\cite{disentangled}
& BiasEnsemble~\cite{BiasEnsemble}

\\
\cmidrule{4-12}
&&
& \textcolor{black}{\boldxmark} \textcolor{black}{\boldxmark}               
& \textcolor{black}{\boldcheckmark} \textcolor{black}{\boldcheckmark}       
& \textcolor{black}{\boldcheckmark} \textcolor{black}{\boldcheckmark}       
& \textcolor{black}{\boldxmark} \textcolor{black}{\boldcheckmark}           
& \textcolor{black}{\boldxmark} \textcolor{black}{\boldcheckmark}           
& \textcolor{black}{\boldxmark} \textcolor{black}{\boldcheckmark}           
& \textcolor{black}{\boldxmark} \textcolor{black}{\boldxmark}               
& \textcolor{black}{\boldxmark} \textcolor{black}{\boldxmark}               
& \textcolor{black}{\boldxmark} \textcolor{black}{\boldxmark}               
\\
\midrule
\multirow{5}{*}{\makecell{Colored \\ MNIST}} 
& 0.5 pct &  & 27.04\stdv{3.12} & 26.99\stdv{5.34} & 29.37\stdv{3.46} & 15.91\stdv{4.59} & 12.56\stdv{2.84} & 23.37\stdv{1.91} & 59.05\stdv{7.60} & 57.98\stdv{1.88}& \underline{\textbf{64.25}}\stdv{2.88}\\ & 1.0 pct &  & 44.11\stdv{0.82} & 40.74\stdv{4.47} & 53.13\stdv{2.18} & 33.29\stdv{13.22} & 23.06\stdv{4.95} & 41.66\stdv{5.02} & 74.88\stdv{1.28} & 72.14\stdv{2.77}& \underline{\textbf{76.54}}\stdv{1.98}\\ & 2.0 pct &  & 63.05\stdv{1.48} & 56.96\stdv{3.49} & 67.48\stdv{0.77} & 46.00\stdv{13.26} & 40.97\stdv{6.45} & 56.37\stdv{2.25} & 81.58\stdv{1.24} & 75.00\stdv{4.67}& \underline{\textbf{82.20}}\stdv{0.84}\\ & 5.0 pct &  & 79.48\stdv{0.82} & 76.20\stdv{1.15} & 79.41\stdv{2.03} & 77.29\stdv{2.92} & 70.25\stdv{2.69} & 77.19\stdv{0.83} & 85.44\stdv{1.45} & 82.46\stdv{0.38}& \underline{\textbf{88.45}}\stdv{0.50}\\ & 20.0 pct & & \underline{\textbf{92.01}}\stdv{0.33} & 89.82\stdv{0.71} & 91.41\stdv{0.89} & 90.07\stdv{0.25} & 90.80\stdv{0.61} & 88.51\stdv{0.42} & 89.95\stdv{0.73} & 91.13\stdv{0.35} & 90.43\stdv{1.71}\\
\midrule
\multirow{5}{*}{\makecell{BFFHQ}} 
& 0.5 pct &  & 48.78\stdv{1.41} & 53.32\stdv{1.58} & 51.62\stdv{2.05} & 52.84\stdv{1.50}& \textbf{54.52}\stdv{4.40} & 53.36\stdv{1.61}& \underline{52.52}\stdv{3.63} & 46.78\stdv{2.29} & 47.94\stdv{4.25}\\ & 1.0 pct &  & 57.26\stdv{1.25} & 57.28\stdv{1.82} & 59.98\stdv{1.61} & 60.28\stdv{2.04} & 63.00\stdv{1.89} & 58.92\stdv{1.23} & 60.72\stdv{2.46} & 58.02\stdv{2.20}& \underline{\textbf{63.14}}\stdv{2.12}\\ & 2.0 pct &  & 64.32\stdv{0.71} & 69.18\stdv{2.48} & 65.90\stdv{2.89} & 67.78\stdv{1.68}& \textbf{73.56}\stdv{1.24} & 68.46\stdv{2.31} & 67.48\stdv{1.83} & 66.06\stdv{3.47}& \underline{71.96}\stdv{4.77}\\ & 5.0 pct & & \underline{80.32}\stdv{2.04} & 82.60\stdv{1.32} & 78.24\stdv{1.72} & 81.54\stdv{1.44}& \textbf{84.98}\stdv{0.62} & 82.52\stdv{2.26} & 69.38\stdv{1.65} & 78.88\stdv{0.96} & 73.36\stdv{1.96}\\ & 20.0 pct & & \underline{92.92}\stdv{0.83} & 93.58\stdv{0.68} & 93.24\stdv{0.82} & 92.78\stdv{0.43}& \textbf{94.44}\stdv{0.52} & 92.98\stdv{0.33} & 81.36\stdv{3.16} & 89.32\stdv{2.41} & 90.48\stdv{1.00}\\
\midrule
\multirow{5}{*}{\makecell{Waterbirds}} 
& 0.5 pct &  & 37.18\stdv{16.71} & 38.39\stdv{3.61} & 40.72\stdv{12.88} & 54.18\stdv{14.55} & 42.63\stdv{8.89}& \textbf{77.69}\stdv{0.65} & 56.00\stdv{4.67} & 56.39\stdv{8.30}& \underline{65.99}\stdv{8.96}\\ & 1.0 pct &  & 37.91\stdv{12.94} & 42.05\stdv{11.42} & 47.15\stdv{17.16} & 71.25\stdv{7.31} & 38.80\stdv{2.95}& \textbf{74.86}\stdv{3.57} & 62.19\stdv{3.80} & 53.93\stdv{3.52}& \underline{64.74}\stdv{7.37}\\ & 2.0 pct &  & 47.48\stdv{7.37} & 41.76\stdv{5.19} & 74.86\stdv{6.61} & 65.50\stdv{17.43} & 44.96\stdv{3.72}& \textbf{76.43}\stdv{1.97}& \underline{75.68}\stdv{2.74} & 37.89\stdv{32.20} & 29.82\stdv{4.18}\\ & 5.0 pct &  & 55.93\stdv{6.83} & 49.56\stdv{5.71} & 64.69\stdv{12.01} & 59.96\stdv{11.53} & 50.79\stdv{5.61}& \textbf{77.52}\stdv{0.78} & 67.78\stdv{5.22}& \underline{71.02}\stdv{4.06} & 52.85\stdv{7.06}\\ & 20.0 pct &  & 69.47\stdv{1.15} & 71.24\stdv{2.05} & 70.71\stdv{2.19} & 71.00\stdv{1.65} & 68.92\stdv{1.84} & 67.97\stdv{3.11}& \underline{\textbf{72.04}}\stdv{3.82} & 70.86\stdv{3.22} & 67.20\stdv{2.55}\\
\midrule
\multirow{3}{*}{\makecell{BAR}} 
& 1.0 pct &  & 64.79\stdv{3.43} & - & - & 63.29\stdv{4.79} & 65.14\stdv{2.30} & 57.14\stdv{6.07} & 65.36\stdv{2.97}& \underline{\textbf{66.50}}\stdv{2.68} & 66.07\stdv{1.18}\\ & 5.0 pct &  & 82.71\stdv{0.74} & - & - & 78.86\stdv{3.49} & 79.36\stdv{1.64} & 74.86\stdv{1.72} & 82.36\stdv{1.25} & 83.79\stdv{1.95}& \underline{\textbf{84.43}}\stdv{0.86}\\ & 20.0 pct &  & 90.51\stdv{0.90} & - & - & 88.38\stdv{0.94} & 88.99\stdv{0.55} & 84.24\stdv{2.15}& \underline{\textbf{91.82}}\stdv{1.40} & 91.62\stdv{1.36} & 91.01\stdv{0.83}\\ 
\midrule
\bottomrule
\end{tabular}
}
\vspace{-0.2cm}
\caption{Image classification accuracy evaluated with only bias-conflicting samples in the unbiased tests sets. The \textit{cross} and \textit{check} marks denote whether each algorithm requires 1) bias labels during training and 2) prior knowledge on bias types.}
\label{supple:main_f1_acc_conflict}
\end{sidewaystable}

\end{document}